# Defaults and Infinitesimals
# Defeasible Inference by Nonarchimedean Entropy Maximization

Emil Weydert
Max-Planck-Institute for Computer Science
Im Stadtwald, D-66123 Saarbrücken, Germany
emil@mpi-sb.mpg.de

## Abstract

We develop a new semantics for defeasible inference based on extended probability measures allowed to take infinitesimal values, on the interpretation of defaults as generalized conditional probability constraints and on a preferred-model implementation of entropy-maximization.

## 1 INTRODUCTION

Probabilistic and default inference, emphasizing either fine- or coarse-grained uncertainty management, scientific or commonsense analysis, constitute the cornerstones of plausible reasoning. Probabilistic approaches are based on a well-established, largely canonical mathematical framework and have a long tradition of successful applications in numerous areas. In addition, they have been recently strengthened by more qualitative accounts allowing an efficient encoding and exploitation of structural information, e.g. about independencies and causal relationships. Default formalisms, on the other hand, have a less impressive record. In fact, after fifteen years of intensive research, the discipline is still characterized by a confusingly large number of competing, partly quite isolated proposals, which are often ad hoc, poorly motivated and frequently unable to produce either answers we would expect or at least explanations why our expectations would be misleading. Nevertheless, there are lots of inferential tasks involving uncertainty - more or less efficiently executed by humans - where standard numerical representations and strategies turn out to be computationally expensive, cumbersome, non-intuitive or even meaningless. So, there is still a strong feeling that we need the concept of a default for practical reasoning and, hopefully, that we should also be able to make explicit our corresponding intuitions.

The most promising approaches - at least as far as foundational coherence and arguably correct behaviour are concerned - appear to be those derived from the probabilistic paradigm, e.g. ranking measures for interpreting default conditionals [Wey 91, 94] and revisable belief strengths [Spo 90], and the limiting-probability accounts for defeasible inference based on minimal information and indifference principles [GMP 90, BGHK 93]. There are several reasons why the quasi-probabilistic perspective is particularly attractive. First, it helps us to avoid serious conflicts between default conclusions and probabilistic judgments. Here, we should keep in mind that probability theory is strongly backed not only by practical considerations, but also by foundational reflections about rational behaviour. Next, it makes it easier to evaluate how reasonable or reliable our defeasible reasoning patterns are, because we obtain a reference point allowing us to identify and clarify the assumptions, simplifications and compromises behind our inference philosophies. A further advantage is the possibility to exploit some powerful classical probabilistic tools. Last but not least, this approach facilitates decision-theoretic investigations, which are the main purpose of plausible reasoning in the real world.

In this paper, we are going to continue this research program and present an integrated framework for extended probabilistic and default reasoning, whose task is to offer a powerful normative background for practical realizations. More precisely, we are going to interpret defaults by generalized probability constraints, which opens the way to the implementation of promising new plausible inference strategies. We begin with a critical look at the ranking measure paradigm for default knowledge and propose to replace it by a more powerful semantic perspective using nonarchimedean probability measures, i.e. distributions allowed to take infinitely small ($\neq$ 0) values. A central idea is to develop defeasible entailment strategies based on extended information-theoretic notions. But to achieve this, we need valuation algebras offering more sophisticated algebraic tools, like exponentiation or logarithmization, as well as some sophisticated results from model-theoretic algebra. On top of these nonstandard measures, we investigate and discuss several possible interpretation policies for defaults. Based on this, we then propose a general four-step-methodology for defeasible inference. In particular, we exploit nonarchimedean entropy maximization to implement the minimal information philosophy for nonmonotonic inference. This amounts to define a suitable preference relation on non-archimedean probability distributions, which allows us to validate Lehmann's postulates for preferential inference. To conclude, we provide a comparison with other popular approaches and illustrate our entailment relations and their competitors by the way they handle relevant benchmark problems.

## 2 RANKING MEASURE SEMANTICS

What is a default ? Generally speaking, it is a binary relationship between propositions, written $\varphi \dashrightarrow \psi$, whose role is to guide plausible inference processes. In practice, accepting a default $\varphi \dashrightarrow \psi$ means that if $\varphi$ were the only fact we knew, then we would be willing to assume or expect that $\psi$ is true as well. In what follows, we are going to adopt the descriptive paradigm, which sees defaults as representing strong conditional expectations effectively anchored in some objective or subjective reality. That is, they are supposed to have global truth conditions and to tell us - in a very simplified manner - something about common or normal relationships in the real world resp. in our epistemic model of it. This reading immediately suggests a quasi-probabilistic semantics for default knowledge, which may be informally described by

- $\varphi \dashrightarrow \psi$ *holds iff* $P(\neg \psi \mid \varphi)$ *is extremely small*.

How should we interpret P and "extreme smallness" ? A first possibility would be to use a coarse-grained form of quasi-probabilistic valuations called ranking measures. This approach was originally advocated in [Wey 91] and further developed in [Wey 94][1].

**Definition 2.1** A function $\mathcal{R} : B \to V$ is called a *ranking measure* iff

1. $\mathcal{B} = (B, \cap, \cup, -, 0, 1)$ is a boolean algebra (e.g. of propositions),
2. $\mathcal{V} = (V, *, \ll)$ is a *ranking algebra* : $(V \setminus \{-\infty\}, *, \ll)$ is the negative half of a nontrivial ordered commutative group[2] with identity o and, for all v∈ V, $-\infty \underline{\ll} v$ («-minimum) and $-\infty * v = v * -\infty = -\infty$ (absorptive for *),
3. For A, A' ∈ B, $\mathcal{R}(A \cup A') = \max_{\ll}\{\mathcal{R}(A), \mathcal{R}(A')\}$ and $\mathcal{R}(0) = -\infty, \mathcal{R}(1) = o$,
4. $\mathcal{R}(A) = -\infty$ if $A = \cup_{\mathcal{B}}\{A_i \mid i \in I\}$ and, for all i∈ I, $\mathcal{R}(A_i) = -\infty$ (*coherence*).

The *conditional ranking measure* corresponding to $\mathcal{R}$ is defined by the equations $\mathcal{R}(A \cap B) = \mathcal{R}(B \mid A) * \mathcal{R}(A)$ for $\mathcal{R}(A) \neq -\infty$, and $\mathcal{R}(B \mid A) = -\infty$ for $\mathcal{R}(A) = -\infty$.

Ranking measure values can be seen either - on the objective side - as coarse, simplifying representations of extreme probabilities in the real world or - on the subjective side - as rough degrees of disbelief or potential surprise, i.e. the smaller $\mathcal{R}(\neg A)$, the stronger our belief in the proposition A. So, $\mathcal{R}(A) \ll \mathcal{R}(B)$ basically means that in some sense, A is negligible w.r.t. B. For us, the main role of ranking measures is to provide a transparent descriptive monotonic semantics for default implication. In fact, extreme smallness can be expressed in a natural way by "$\ll$ o". Let L be a first-order language and $\mathcal{B}_L = (B_L, \&, \vee, \neg, F, T)$ be the corresponding (compact) boolean Lindenbaum algebra[3] induced by first-order predicate logic. The standard ranking measure semantics for defaults is now easily given (for $\varphi, \psi \in L$) by

- $\mathcal{R}$ *satisfies* $\varphi \dashrightarrow \psi$ iff $\mathcal{R}(\neg\psi \mid \varphi) \ll$ o
  iff $\mathcal{R}(\varphi \& \neg\psi) \ll \mathcal{R}(\varphi \& \psi)$ or $\mathcal{R}(\varphi \& \neg\psi) = -\infty$.

For this interpretation, a sound and complete axiomatization of default conditionals is offered by (an object-level, boolean version of) Lehmann's rational conditional logic [Wey 91]. We could now strengthen our default concept by imposing more restrictive bounds on the corresponding conditional ranking measures, e.g. $\mathcal{R}(\neg\psi \mid \varphi) \ll$ e or $\mathcal{R}(\neg\psi \mid \varphi) \underline{\ll}$ e for e « o. It is not difficult to see that under this reading, the rules of preferential conditional logic [KLM 90] are still guaranteed. However, rational monotony ($\alpha \dashrightarrow \beta$ and not $\alpha \dashrightarrow \neg\alpha'$ implies $\alpha \& \alpha' \dashrightarrow \beta$) is no longer valid, as the following example shows. Suppose, $\mathcal{R}$ is a Spohn-type κ-ranking, i.e. a ranking measure whose ranking algebra is based on the additive structure of negative integers ($\{-\infty, \ldots -2, -1, 0\}, +, <$) and which verifies $\mathcal{R}(\neg\varphi) = -2, \mathcal{R}(\neg\varphi \& \psi) = -2$ and $\mathcal{R}(\varphi \& \psi) = -1$. Interpreting $\alpha \dashrightarrow \beta$ by $\mathcal{R}(\neg\beta \mid \alpha) \ll -1$ ($\ll$ 0), we then get $T \dashrightarrow \varphi$, but neither $T \dashrightarrow \neg\psi$ nor $\psi \dashrightarrow \varphi$, because $\mathcal{R}(\neg\varphi \mid T) = -2$, $\mathcal{R}(\psi \mid T) = -1$ and $\mathcal{R}(\neg\varphi \mid \psi) = -1$. But, there are also more basic problems affecting the ranking measure approach.

**1.** It is too coarse-grained for most decision-theoretic purposes. For instance, we cannot make a difference between a situation where we have gotten 100 positive and 1 negative equally plausible outcomes and one where the the odds are inversed. The main reason is that equal ranking measure values don't add up.

**2.** It is somewhat isolated from standard probabilistic approaches, because its valuation structures are too rudimentary. This means that there are only indirect possibilities to exchange informations and powerful methods. Furthermore, it is difficult to give a precise, objective meaning to ranking measure values. On the other hand, a direct connection to or a common ground with the fine-grained world of real probability could well provide a better theoretical understanding making it easier to motivate, develop and investigate more reasonable quasi-probabilistic plausible consequence relations for default knowledge.

However, we should not forget that the classical probabilistic approach itself has notorious defects. For instance, it may be inappropriate when precise numbers are not required, unavailable or meaningless and the computational costs become disproportionate w.r.t. to the inferential needs. This is and was a major justification for the switch to ranking measures. But probability measures are also haunted by technical and foundational weaknesses. A very prominent one is the non-existence of uniform distributions on infinite sets as long as we want to ban non-empty sets with vanishing probability, reserving measure zero for real impossibility. The difficulties to represent revisable - i.e. belief negations with non-zero probability - plain belief - i.e. closed under conjunction and logical consequence - in a natural way by classical subjective probabilities, interfer with the desire for dynamic logical accounts [Spo 90]. Standard probabilistic threshold interpretations for defaults are easily seen to be absolutely unsuitable. Translating $\varphi \twoheadrightarrow \psi$ by $P(\neg\psi \mid \varphi) \leq \alpha$, for non-zero $\alpha$, would be completely ad hoc and in flagrant conflict with our intuitions. But $P(\neg\psi \mid \varphi) \leq 0$ is inacceptable as well because it would necessarily induce the trivialization of defaults with an exceptional antecedent, e.g. automatically validating $\varphi \& \neg\psi \twoheadrightarrow \neg\varphi$ and $\varphi \& \neg\psi \twoheadrightarrow \varphi$. So, there are several reasons to look for an extended framework.

---

[1] It turns out to generalize Spohn's natural conditional functions [Spo 90], i.e. Pearl's κ-rankings, and Dubois and Prade's [DP 88] possibility measures.

[2] (G, *, «) is an ordered commutative group iff * is associative, commutative, has a neutral element, admits inverses and « is a strict total order satisfying x « y -> x*v « y*v.

[3] For the sake of notational economy, we sloppily denote the elements of $B_L$ (the sets of classically equivalent L-formulas) by their representatives from L.

## 3 NONARCHIMEDEAN PROBABILITIES

We have seen that the traditional probabilistic as well as related coarse-grained approaches both fail - for different reasons - to meet our expectations. The major remaining alternative then is to consider generalized probability measures using more fine-grained valuation algebras extending ([0, 1], +, x, <). Because we want to conserve as much of the classical probabilistic context as possible, the most interesting candidates are finitely additive measures taking their values from the unit interval of some ordered field[4] $IR' = (R', 0, 1, +', x', <')$ properly extending the standard ordered real number field $IR = (R, 0, 1, +, x, <)$, ie. $R' \supseteq R$ s.t. the restrictions of $+'$, $x'$, $<'$ to $R$ are just $+$, $x$, $<$[5]. First, note that such an $IR'$ is necessarily a nonarchimedean extension of $IR$, i.e. it includes $\varepsilon \neq 0$ with $-r < \varepsilon < r$ for every strictly positive standard real $r$. These $\varepsilon$ are called infinitesimals. To deal with them, we define for $0 \leq a, b \in R'$ the much-smaller-than-ordering $\ll$. $a \ll b$ iff $a < r \times b$ for all standard reals $r > 0$. This makes $\ll$ a strict total pre-order on the positive half of $R'$. Because we are going to need an extended logarithm function for defining information measures aimed at nonarchimedean probability distributions, in fact, we have to look for proper extensions $IR'$ of the real ordered exponential field $IR_e = (R, 0, 1, +, x, exp_2, <)$, where $exp_2(r)$ stands for $2^r$. Let $L_{EF}$ be the first-order language of ordered exponential fields and $L_{EF}(R)$ be the expansion of $L_{EF}$ by constants $\underline{r}$ for each real number $r$. The existence of such $IR'$ is now guaranteed by the compactness theorem ($\{\varphi \mid IR_e \models \varphi, \varphi \in L_{EF}(R)\} \cup \{0 < x, x < \underline{r} \mid r \in R\}$ is finitely satisfiable over $R_e$, hence satisfiable). There exists a correct, presumably incomplete axiomatization RCE of the $L_{EF}$-theory $Th(IR_e)$ of the authentic real ordered exponential field.

**Definition 3.1** *Let* **RCF** *be the theory of real closed ordered fields, i.e. ordered fields where every polynomial of odd degree has a root (= theory of* $IR$*). The structure* $IF = (F, 0, 1, +, x, exp, <)$ *is called an ordered real closed exponential field iff* $IF$ *verifies* $RCE = RCF + E1 - E4$,

**E1** $\exp(1) = 1+1, \exp(v + w) = \exp(v) \times \exp(w)$,
**E2** $v < w \rightarrow \exp(v) < \exp(w)$,
**E3** $0 < v \rightarrow \exists w \exp(w) = v$,
**E4** $\underline{n}^2 < v \rightarrow v^n < \exp(v)$, for $n \in Nat$ ($v^n = v \times ... \times v$).

*The corresponding logarithmic function log is defined by* $\log(v) = w$ *iff* ($v \leq 0$ *and* $w = 0$) *or* ($\exp(w) = v$).

Of course, for practical and foundational reasons, $IR'$ should be as close to $IR_e$ as possible. A straightforward way to achieve this is to require that $IR'$ satisfies $Th(IR_e)$. Among others, we may then freely use classically definable notions. But we can go even one step further. First, some definitions. An L-theory T is called *model-complete* iff for any pair of T-models M and N, if N is an extension of M, then N is an elementary extension of M, i.e. every L-sentence with parameters from M is true in N iff it holds in M. In other words, as far as first-order logic is concerned, elementary extensions don't bring anything new. Another equivalent way to put it is to say that $T \cup \{\varphi \mid M \models \varphi, \varphi \text{ quantifier-free } L(M)\text{-formula}\}$ is complete in $L(M)$, i.e. L extended by constants for the elements of M. Now Wilkie has shown the following [Dri 94].

**Proposition 3.1** $Th(IR_e)$ *is model-complete.*

That is, by assuming that our $IR_e$-extension $IR'$ validates $Th(IR_e)$, we can ensure that every classical result about $IR_e$, obtained by any means but expressible in $L_{EF}(R)$, also holds in $IR'$. In other words, that $IR'$ is an elementary extension of $R_e$. But Wilkie was also able to prove that $Th(IR_e)$ is o-minimal, which means that the only parameter-definable sets $\{a \in F \mid IF \models \varphi(\underline{a}), \varphi(x) \in L_{EF}\}$ in its models $IF$ are finite unions of open intervals and points. From this it follows that all the strictly positive infinitesimals in $IR'$ satisfy the same $L_{EF}(R)$-formulas with a single free variable. In fact, unary properties in the realm of smallness are completely fixed by the limit-behaviour of the standard reals.

**Proposition 3.2** *Let* $IR' = (R', 0, 1, +, x, exp, <)$ *be a proper extension of* $IR_e$ *satisfying* $Th(IR_e)$ *and* $\varepsilon, \tau > 0$ *be infinitesimals in* $IR'$. *Then, for all* $\varphi(x) \in L_{EF}(R)$, $IR' \models \varphi(\varepsilon)$ *iff there is* $0 < r \in R'$ *s.t. for all* $s \in R'$, $0 < s < r$ *implies* $IR' \models \varphi(\underline{s})$ *iff there is* $0 < r \in R$ *s.t. for all* $s \in R$, $0 < s < r$ *implies* $IR \models \varphi(\underline{s})$ *iff* $IR' \models \varphi(\tau)$.

This means in particular that there are no privileged non-zero infinitesimal values, which corresponds well to our intuitions about the indiscernibility of infinitesimals. We cannot differentiate infinitely small numbers independently from other nonstandard reference points. Hence, each infinitesimal might be called prototypical or generic in the context of Th($\mathbb{R}_e$). Observe that all this would not be true if we could refer to an integer concept allowing the distinction of odd- and evenness. We are now ready to provide a more formal definition of our extended probability measures.

---
[4] An axiom set for ordered fields can be found in [Bac 90].

[5] To ease notation, from now on, we are going to use the same relational/functional notation for structures and their extensions.

**Definition 3.2** *Let* $\mathbb{R}'$ *be a* Th($\mathbb{R}_e$)-*model extending* $\mathbb{R}_e$ *and* $\mathfrak{B} = (B, \cap, \cup, \neg, 0, 1)$ *be a boolean algebra. The function* $P : B \to \mathbb{R}'$ *is called an* $\mathbb{R}'$-*valued nonarchimedean probability measure iff for all* A, B$\in$B, P(A) $\geq$ 0, P(1) = 1 *and* P(A$\cup$B) = P(A) + P(B) *for disjoint* A, B. *It is called coherent if* P(A) = 0 *only holds for* A = **0**.

Suppose, P is an $\mathbb{R}'$-valued probability measure on B, and $\{A_i \mid i \in \text{Nat}\}$ a subset of B s.t. $A_i \cap A_j = \mathbf{0}$ for $i \neq j$, $P(A_i) = 1/2^{i+1}$ and $A = \cup_\mathfrak{B} A_i$ (unique least upper bound w.r.t. $\mathfrak{B}$). But then, because the infinite sum of the $P(A_i)$ is just defined to be the supremum of the partial sums $s_i = 1/2 + ... + 1/2^{i+1}$, which doesn't exist in $\mathbb{R}'$ (for any potential limit s, we would get $s_i < s - \epsilon < s$ for all i), we cannot set $P(A) = \Sigma P(A_i)$, i.e. we cannot assume $\sigma$-additivity. Coherence, i.e. respecting impossibility, can only be a facultative requirement given that important classical distributions like the Lebesgue measure on [0, 1] violate this principle. In practice, however, we are mostly concerned with restrictions of $\mathbb{R}'$-valued measures to boolean algebras of definable sets or Lindenbaum-algebras $\mathfrak{B}_L$. Note that we may construct from a given coherent nonarchimedean probability measure $P : B \to [0, 1]_{\mathbb{R}'}$ a corresponding canonical ranking measure $\mathfrak{R}^P$ by identifying those values a, b in $[0, 1]_{\mathbb{R}'}$ which are <<-incomparable (not a << b or b << a) and taking the quotient structure. The ranking algebra resulting from this construction will be dense, contrasting with the discrete $\kappa$-ranking structures. We get the following extendibility result.

**Proposition 3.3** *Let* $\mathfrak{B}$ *be a boolean subalgebra of the power-set-algebra* $\mathcal{P}ow(S)$, $\mathbb{R}'$ *be as above and* P *be an* $\mathbb{R}'$-*valued probability measure on* B. *Then there is a* Th($\mathbb{R}_e$)-*model* $\mathbb{R}''$ *extending* $\mathbb{R}'$ *and an* $\mathbb{R}''$-*valued probability measure* P' *on* Pow(S) *extending* P.

What is the real meaning of infinitesimal probabilities ? Certainly, it seems hard to imagine any objective, statistical interpretation, in particular because of our indiscernibility result. That is, we have to adopt a subjectivist perspective. Then, their role is three-fold. To begin with, they provide a rough classification of propositions according to their respective relevance or order of magnitude. For an arbitrary infinitesimal $\epsilon$, $P(A \mid B) = \epsilon$ practically means that within the context B - at least for finite boolean algebras - we should neglect the alternative A for utility considerations and decision-taking. Similar to ranking measure values, they allow us to express absolute and relative negligibility. Of course, this could be done as well through $\mathfrak{R}^P$. The fine-grained distinctions offered by the extended probabilistic scale matter when extreme conditioning on neglected evidence has to occur and we have to revise probabilities (e.g. to prepare a decision). So, their second task is to supply a coherent framework for borderline conditionalizations and more subtle comparisons within the same order of magnitude. Last but not least, we are going to use them in the following for more fine-grained interpretations of default knowledge and reasoning, trying to accommodate established probabilistic inference mechanisms to our extended framework and thereby to defeasible inference.

To conclude, some words about previous uses of nonstandard probabilities in the context of default reasoning and belief revision. Spohn [Spo 90] exploited this notion to interpret and justify his discrete $\kappa$-ranking revision framework. He did so by associating $\kappa(A) = i$ with $P(A) = \epsilon^i$ for an arbitrary but fixed infinitesimal $\epsilon$. Nonstandard valuations admitting infinitesimals were also used in [LM 92] for a probabilistic reinterpretation of ranked models. A difference with our approach is that they worked within the traditional formal context of nonstandard analysis [Sto 77]. That is, they extended not only the real number field but also the corresponding set theory, which is much more than we need and want to do regarding our intuitions and intentions. We make only those assumptions necessary to model defaults and defeasible inference in a more fine-grained probabilistic context. The adequacy of our way to proceed is guaranteed by the previously mentioned model theoretic results. There have been other proposals combining the probabilistic with the ranking measure perspective, e.g. counterfactual probabilities [Bou 93] and cumulative measures [Wey 94], but these heterogeneous approaches are sometimes less adequate. For practical purposes, of course, they may still constitute a real alternative.

## 4 DEFAULT KNOWLEDGE

Infinitesimal probabilities are certainly a very natural way to encode the notion of "negligibility" or "extreme smallness", which is at the center of our descriptive default philosophy. However, interpreting defaults by nonarchimedean conditional probability constraints is not entirely straightforward. The reason is that we have to deal with a multitude of possible interpretations, which may produce different results. In what follows, let L be a classical propositional or first-order language and P be a nonarchimedean $\mathbb{R}'$-valued probability measure on the corresponding Lindenbaum algebra $\mathfrak{B}_L$. To implement the attribute extreme smallness, a subset of $[0, 1]_{\mathbb{R}'}$ must satisfy at least two conditions, namely downwards closure and infinitesimality. Therefore, in our framework, the translation of a default $\varphi \twoheadrightarrow \psi$ ($\varphi, \psi$ L-formulas) will always take the form of an infinitesimal positive initial interval constraint,

- $\varphi \twoheadrightarrow \psi :$ $P(\neg\psi \mid \varphi) \in I_{\varphi \to \psi}$, with

$[0, 1]_{\mathbb{R}'} \supseteq I_{\varphi \to \psi} \neq \emptyset$, s.t. for all $v \in I_{\varphi \to \psi}$, $0 \leq v \ll 1$
and $0 \leq w \leq v$ implies $w \in I_{\varphi \to \psi}$.

Observe that we allow our intervals to depend on $\varphi \twoheadrightarrow \psi$. To define $I_{\varphi \to \psi}$, we may now choose between different comparison relations (<, $\leq$, <<, $\leq\leq$) and bounds (infinitesimal terms, 1). But, what are the most appropriate strategies for translating the descriptive content of defaults by such constraints ? Here, we have to take into account not only our intuitions about the intended logical behaviour of defaults, e.g. w.r.t. Lehmann's conditional axioms, but also the possible defeasible inference policies which will have to exploit this encoding of default knowledge. An obvious choice would be to use the canonical ranking measure construction and to transfer the corresponding default interpretations. This would give us three kinds of constraints based on the magnitude relation <<. Let $\varepsilon_{\varphi \to \psi}$ << 1. Note that $\not\ll$ stands for not >>.

- $\varphi \to \psi$ : $P(\neg\psi \mid \varphi) \ll 1$ (<<-classical),
- $\varphi \to \psi$ : $P(\neg\psi \mid \varphi) \ll \varepsilon_{\varphi \to \psi}$ (<<-bounded),
- $\varphi \to \psi$ : $P(\neg\psi \mid \varphi) \not\gg \varepsilon_{\varphi \to \psi}$ ($\not\gg$-bounded).

Recalling what we said about ranking measures and the fact that $\varepsilon \ll \sqrt{\varepsilon} \ll 1$ for $\varepsilon \ll 1$, we see that the first scheme guarantees the rules of rational, the second and the third one only those of preferential conditional logic. So, only the <<-classical interpretation shows the desired behaviour at the level of monotonic inference for defaults. The corresponding proof theory then allows us to transform given default knowledge bases without changing their intended nonarchimedean probabilistic meaning. The precise syntactic form doesn't matter under these conditions. Most old-fashioned default formalisms, like Reiter's default logic, do not support such natural, content-preserving manipulations. To simplify the overall valuation context, we may try to restrict the values P can take, e.g. to polynomials over a fixed infinitesimal $\varepsilon$. In the context of ranking measures, this would correspond to the choice of a discrete ranking algebra.

Another possibility would be to interpret all our defaults through infinitesimal, but otherwise classical threshold constraints using < or ≤.

- $\varphi \to \psi$ : $P(\neg\psi \mid \varphi) < \varepsilon_{\varphi \to \psi}$ (<-bounded),
- $\varphi \to \psi$ : $P(\neg\psi \mid \varphi) \leq \varepsilon_{\varphi \to \psi}$ (≤-bounded).

These constraints, however, strongly conflict with almost all the classical postulates for default conditionals (e.g. conjunction on the right, reasoning by cases, cautious monotony). Nevertheless, that doesn't make this interpretation worthless. For instance, ≤-boundedness is appropriate for implementing defeasible inference strategies based on generalized probabilistic techniques like nonarchimedean entropy maximization. Furthermore, on an intuitive level, this interpretation and the <<-classical one are quite close to each other. In fact, asking $P(\neg\psi \mid \varphi)$ to be << 1, i.e. to be ≤ than some (generic) infinitesimal, doesn't seem to be so different from asking it to be ≤ than a fixed (generic) infinitesimal, given the indiscernibility result from Prop. 3.2. Of course, this argument breaks down if we consider not one but several infinitesimal bounds in parallel (e.g. $P(\neg\psi \mid \varphi) \leq \varepsilon$ and $P(\neg\psi' \mid \varphi') \leq \varepsilon' = \varepsilon^2$). So, it is not completely obvious which translation strategy serves our needs best, because there seems to be a surprising trade-off between having a suitable monotonic logic for defaults and getting nice nonmonotonic inference relations to exploit them.

## 5 DEFAULT INFERENCE

In its traditional form, defeasible reasoning has been mainly concerned with finite splitted knowledge bases of the type $\Sigma \cup \Delta$, where $\Sigma = \{\varphi_i \mid i \leq n_\Sigma\}$ is a set of facts from L and $\Delta = \{\psi_i \to \psi'_i \mid i \leq n_\Delta\}$ a collection of defaults over L, whose role is to guide a plausible reasoning process starting at $\Sigma$. The task of any - necessarily nonmonotonic - plausible inference relation $\Vdash$ is to tell us, given facts and defaults, what we might reasonably expect or assume beyond certainty, i.e. which $\psi$ should be defeasibly inferred (*is a plausible consequence of* $\Sigma \cup \Delta$).

- $\{\varphi_i \mid i \leq n_\Sigma\} \cup \{\psi_i \to \psi'_i \mid i \leq n_\Delta\} \Vdash \psi$.

For each $\Delta$, we can then define a finitary consequence relation $\vDash_\Delta$ on L by setting $\Sigma \vDash_\Delta \psi$ iff $\Sigma \cup \Delta \Vdash \psi$. Currently, a rather broad consensus has been reached about the minimal requirements for such factual inference relations $\vDash_\Delta$ (but not for interactions between defaults). They are at least required to be *preferential* for finite premise sets [KLM 90]. We do not ask for rational monotony, because sometimes, we may want to construct new inference relations by taking the intersection of existing ones and to see $\vDash_\Delta$ as giving only a partial approximation of some ideal plausible relationships, which conflicts with the above principle. Concerning the full relation $\Vdash$ and the impact of $\Delta$, however, there doesn't exist a general agreement about which postulates to adopt, in particular if the basic default notion is allowed to violate the rules of preferential conditional logic. For us, the primary criterion is the existence of a semantic justification which doesn't conflict with the probabilistic perspective and might even be justified by probabilistic considerations. So, let's turn now to default inference in the context of our nonarchimedean framework of $\mathbb{R}'$-valued probability measures.

The basic idea governing the defeasible inference philosophy we are going to adopt here is that

*Default reasoning should be anchored - on an abstract, ideal level - in the comparison of nonarchimedean probabilistic belief valuations ordered according to some suitable version of the minimal informational commitment principle.*

More precisely, we adopt the following four-step strategy.

**1. Finiteness.** Fix a background language L, closed under the usual propositional connectives, and a corresponding monotonic inference relation $\vdash$, closed under the rules of propositional logic, s.t. $\vdash$ induces on L a finite Lindenbaum algebra $\mathfrak{B}_L$ ($\mathfrak{B}_L$ being the quotient of L over logical equivalence -$\Vdash$-). This should be enough for most practical representational purposes. We may then restrict ourselves to finitary measures, which greatly simplifies matters.

**2. Nonarchimedean constraints.** Translate a given finite default collection $\Delta$ into a set of infinitesimal conditional probability constraints. Several important choices have to be made in this context. First of all, if explicit default strengths or preferences have not been given, we must decide whether our default translation procedure $\mathcal{T}$ should fix a single infinitesimal bound for all the defaults or introduce a particular one for each individual default. Next, we have to choose suitable numerical comparison relations linking bounds and admissible values.

- $\mathcal{T}: \Delta = \{\psi_i \to \psi'_i \mid i \leq n_\Delta\} \to \{P(\neg\psi'_i \mid \psi_i) \in I_i \mid i \leq n_\Delta\}$

The main question here is whether the constraint areas $I_i$ should be closed under addition (requiring $\ll$). If not, the intervals should at least be closed w.r.t. the order-topology induced by $<$ (requiring $\leq$). This gives us three major interpretation strategies $\mathcal{T}$ for $\Delta$, resulting in

- $\Delta^{\mathcal{T}_{\ll 1}} = \{P(\neg\psi'_i \mid \psi_i) \ll 1 \mid i \leq n_\Delta\}$,
- $\Delta^{\mathcal{T}_{\leq,s}} = \{P(\neg\psi'_i \mid \psi_i) \leq \varepsilon \mid i \leq n_\Delta\}$ for $\varepsilon \ll 1$,
- $\Delta^{\mathcal{T}_{\leq,p}} = \{P(\neg\psi'_i \mid \psi_i) \leq \varepsilon_i \mid i \leq n_\Delta\}$ for $\varepsilon_i \ll 1$.

Note that the constraints from the last two sets are open schemes with infinitesimal parameters $\varepsilon$, $\varepsilon_i$. We call $\mathcal{T}_{\ll 1}$ the *classical-bounded*, $\mathcal{T}_{\leq,s}$ the *single-bounded* and $\mathcal{T}_{\leq,p}$ the *plural-bounded* nonarchimedean probabilistic default translation policy.

**3. Preferential entropy reasoning.** Implement the minimal information principle within a preferential framework. This is achieved by defining a partial order on finitary **IR'**-valued probability distributions which gives priority to informationally less committed valuations, i.e. to those making the least additional assumptions given the knowledge at hand. To this end, we have to choose an appropriate information measure. The standard proposal in the literature is the entropy concept. Entropy is in fact a privileged measure for the lack of information which can be justified by uniqueness results for different sets of postulates. Similarly, entropy maximization is a distinguished method in probabilistic inference [Jay 78, PV 90]. Because the axioms for **RCE**-fields already guarantee the existence of a logarithm function corresponding to exp, we may directly use the classical entropy definition in **IR'** as well. Let P be a nonarchimedean **IR'**-valued probability distribution on a finite boolean algebra $\mathfrak{B}$ with atoms $A_1, ..., A_n$. Then the (**IR'**-)*entropy* $H(P)$ of P is defined by

- $H(P) = -\Sigma s_i$ where $s_i = P(A_i) \times \log(P(A_i))$ if $P(A_i) \neq 0$ and $s_i = 0$ if $P(A_i) = 0$ (limit value).

$H(P)$ takes its maximum $\log(n)$ for the uniform distribution $P^\circ$ on $\mathfrak{B}$ (with $P^\circ(A_i) = 1/n$). Note that $H$ is invariant under atom permutations. For the standard valuation structure, it is well-known that every set of linear $\leq$- constraints $\{\Sigma P(A_i)a_{ji} \leq b_j, 0 \leq P(A_i), \Sigma P(A_i) = 1 \mid 1 \leq i \leq n, j \in J\}$ over $(P(A_1), ..., P(A_n))$, which determines a convex subset of $[0, 1]^n$, has a unique entropy-maximizing solution. But note that we may think of other interesting conditions, e.g. concerned with independence, which cannot be expressed by linear means and may require at least polynomial expressions. Now, because **IR'** is an elementary extension of **IR**$_e$, all the $L_{EF}(R)$-expressible results about entropy-maximization in classical finitary contexts, can be immediately transferred to **IR'**, where they hold as well. Based on $H$, we can then define our entropy-based *information-ordering* $<_E$, a total pre-order on finitary **IR'**-valued probability distributions $\bar{P}: B \to [0, 1]_{R'}$.

- $P <_E P'$ iff $H(P') < H(P)$

It is easy to see that the constraint sets $\Delta^{\mathcal{T}}$ provided by the plural- or single-bounded translation policy can be written as sets of linear weak inequality constraints over the atom probabilities $P(A_i)$. Therefore, assuming satisfiability, there exists a unique $<_E$-minimal **IR'**-valued distribution $P^*$ over $\mathfrak{B}$ verifying $\Delta^{\mathcal{T}}$. For constraints based on $<$, $\ll$ or $\leq\leq$, however, there might be no $<_E$-minima at all. Fortunately, we may still use the partial order $<_E$ to define a preferential consequence relation $\Vdash\approx_E$ on all $L_{EF}(R')$-sentences with constants $P(A_i)$. All we have to do - because $<_E$ is neither well-ordered nor otherwise well-behaved - is to adopt the limit evaluation strategy. That is, if $\mathcal{PB}$ is the set of **IR'**-valued distributions over $\mathfrak{B}$,

- $\Phi(P(A_1), ..., P(A_n)) \Vdash\approx_E \psi(P(A_1), ..., P(A_n))$ iff

  in $\mathcal{PB}$, for every P validating $\Phi$, there is a $P' \leq_E P$

  verifying $\Phi$, s.t. for all $P'' \leq_E P'$ satisfying $\Phi$, $\psi$ holds.

**4. Defeasible entailment relations.** Define for every interpretation strategy $\mathcal{T}$ a corresponding defeasible entailment notion $\Vdash\approx^{\mathcal{T}}$ based on $\Vdash\approx_E$, i.e. entropy maximization in nonarchimedean contexts. The general idea is to determine the relationship between a finite premise set $\Sigma$, represented by the conjunction of its elements $\varphi$, and a potential plausible conclusion $\psi$ by evaluating the conditional probability $P(\neg\psi \mid \varphi)$ for the $<_E$-most-preferred distributions verifying $\Delta^{\mathcal{T}}$. The nonmonotonic inference step should succeed iff $P(\neg\psi \mid \varphi)$ is extremely small for all these P, i.e. $P(\neg\psi \mid \varphi) \ll 1$. This guarantees that $\Vdash\approx^{\mathcal{T}}$ is preferential for fixed $\Delta$. Rational monotony cannot be required because usually, $\Sigma$ offers only a partial (defeasible) description of P. Given that we may not omit any possible instan-tiation of the free unspecified infinitesimal parameters $\varepsilon_i$ ($i \leq n_\Delta$) in $\Delta^{\mathcal{T}}$, we are going to accept only defeasible conclusions which are independent from the exact values. To realize this, we proceed by universal quantification. Now, we are ready for our general defeasible entailment scheme. Let $\Sigma = \{\varphi_i \mid i \leq n_\Sigma\}$, $\Delta = \{\psi_i \to \psi'_i \mid i \leq n_\Delta\}$ and $\varphi = \varphi_0 \& ... \& \varphi_{n_\Sigma}$. Then,

- $\Sigma \cup \Delta \Vdash\approx^{\mathcal{T}} \psi$ iff for all $\varepsilon_0, ..., \varepsilon_{n_\Sigma} \ll 1$,

  $\Delta^{\mathcal{T}}(\varepsilon_0, ..., \varepsilon_{n_\Sigma}) \Vdash\approx_E P(\neg\psi \mid \varphi) \ll 1$.

It immediately follows from this definition that $\Vdash\approx^{\mathcal{T}}$ is finitarily preferential for fixed $\Delta^{\mathcal{T}}$, for $\mathcal{T} = \mathcal{T}_{\leq,s}, \mathcal{T}_{\leq,p}, \mathcal{T}_{\ll 1}$. In this paper, we have and will mainly consider standard finite default sets. But, it is important to note that our entailment strategy is in fact applicable to arbitrary constraint sets over nonarchimedean probability measures on some given finite boolean algebra. Furthermore, if we are willing to accept a technically slightly more demanding framework, it is even possible to handle infinite boolean algebras. For practical (normative) purposes, however, fixed finite contexts seem to be sufficient.

# 6 EXAMPLES AND COMPARISONS

To become more acquainted with our nonarchimedean entropy-maximizing entailment strategies, we are now going to illustrate the core approaches $\Vdash\approx_{\leq,p}$, $\Vdash\approx_{\leq,s}$ and $\Vdash\approx_{\ll 1}$ by their impact on some relevant inference patterns and - roughly - investigate their links with other popular accounts from the literature. Notably preferential $\Vdash\approx_{PC}$ and rational closure $\Vdash\approx_{RC}$ [LM 92] (equivalently, for finite sets of conditionals, system **Z** [Pea 90] or elementary hyperentailment [Wey 93, 95]), lexicographic closure $\Vdash\approx_{LC}$ [Leh 92, BCDLP 93], conditional entailment $\Vdash\approx_{CE}$ [GP 92], maximum-entropy entailment $\Vdash\approx_{ME}$ [GMP 90, Gol 92] and random-worlds entailment $\Vdash\approx_{RW}$ [BGHK 93]. The maximum-entropy-based formalism for defeasible reasoning investigated by Goldszmidt is - contrasting with ours - mainly defined for simple default sets interpreted according to $T_{\leq,s}$ and based on a slightly more cumbersome limiting probability strategy. Nonetheless, it gives the same results as $\Vdash\approx_{\leq,s}$ for standard default knowledge bases. This readily follows from Prop. 3.2. A similar remark holds for the random-worlds approach of [BGHK 93], which in fact intends to handle more powerful first-order languages. It combines a somewhat limited approximation methodology with a limiting counting strategy based on very far-reaching indifference assumptions for finite first-order models. This causes difficulties with premises lacking finite models. Last but not least, it should be mentioned that the consequence relation $\Vdash\approx_{\ll 1}$, which is based on the most liberal interpretation strategy $T_{\ll 1}$, turns out to be equivalent to preferential closure $\Vdash\approx_{PC}$. Because the constraint interval here is just the set of all positive infinitesimals (in the given $\mathbb{R}'$), there is a lot of variability when doing entropy-maximization. That's why only the basic $\ll$-relationships restrict the set of admissible nonarchimedean probability distributions (through the rules of probability theory and the definition of $\ll$).

To begin with, we describe several - in our eyes desirable - strict and defeasible (non-)inference features. This list is, of course, not intended to be exhaustive or even representative in a precise sense. It only collects some interesting principles helping us to discriminate, situate and evaluate plausible consequence relations, providing at least a survey of the relevant issues. In the following, we shall assume that $\alpha, \beta, \gamma, \varphi, \psi$ are logically independent formulas from L. Also, we concentrate on standard finite default sets $\Delta$.

**Rat° :** *Object-level rationality.*

- $\rightarrow$» satisfies the rules of rational conditional logic.

These are intuitively appealing for normal implication.

**Pre :** *Meta-level preferentiality.*

- For constant $\Delta$, $\Vdash\approx$ defines a preferential consequence relation on facts.

Fundamental requirement for factual plausible inference.

**Rat :** *Meta-level rationality.*

- For constant $\Delta$, $\Vdash\approx$ defines a rational consequence relation on facts.

At the meta-level, rational monotony is not really required, but sometimes practical.

**ES :** *Extended specificity.*

- $\{\varphi\} \cup \{\varphi \rightarrow\!\!\!» \alpha, \alpha \rightarrow\!\!\!» \beta, \beta \rightarrow\!\!\!» \psi, \alpha \rightarrow\!\!\!»\neg\psi\} \Vdash\approx \neg\psi$.

Extended version of the most-specific-subclass principle, a basic postulate for plausible reasoning with defaults.

**EI :** *Inheritance through exceptional subclasses.*

- $\{\alpha\} \cup \{\alpha \rightarrow\!\!\!» \beta, \alpha \rightarrow\!\!\!»\neg\psi, \beta \rightarrow\!\!\!» \psi, \beta \rightarrow\!\!\!» \varphi\} \Vdash\approx \varphi$.

The formalism should be able to handle implicit independence assumptions coherently. The exceptionality of $\alpha$ in $\beta$ should not affect the inheritance of $\varphi$.

**GE :** *Geffner's example.*

- $\{\alpha\&\beta\&\gamma\} \cup \{\alpha \rightarrow\!\!\!»\neg\psi, \gamma \rightarrow\!\!\!»\neg\psi, \alpha\&\beta \rightarrow\!\!\!» \psi\} \Vdash\!\!/\!\approx \psi, \neg\psi$.

Because $\alpha\&\beta$ and $\gamma$ are unrelated, there is no reason to prefer $\psi$ or $\neg\psi$, given $\alpha\&\beta\&\gamma$.

**AP :** *Anti-prioritization.*

- $\{\alpha\vee(\beta\&\gamma)\} \cup \{T \rightarrow\!\!\!»\neg\beta, T \rightarrow\!\!\!»\neg\gamma, \alpha\vee\beta \rightarrow\!\!\!»\neg\alpha\} \Vdash\!\!/\!\approx \neg\alpha$.

If $\beta$ and $\gamma$ are negligible w.r.t. T and $\alpha$ w.r.t. $\beta$, then expecting $\alpha$ to be negligible w.r.t. $\beta\&\gamma$ seems completely unjustifiable given the presumable exceptionality of $\beta\&\gamma$ w.r.t. $\beta$ (cf. **EI**). Recall that A $\rightarrow\!\!\!»$ B $\approx$ "A&$\neg$B is negligible w.r.t. A(&B)".

**RE :** *Redundancy.*

- $\{\neg\varphi\} \cup \{T \rightarrow\!\!\!» \varphi, T \rightarrow\!\!\!» \varphi\vee\psi\} \Vdash\!\!/\!\approx \psi$.

Assuming right weakening for $\rightarrow$», T $\rightarrow\!\!\!» \varphi\vee\psi$ would become redundant, so we should infer nothing which does not already follow from $\neg\varphi$ and T $\rightarrow\!\!\!» \varphi$ ("syntax-independency").

**NE :** *Neutralization.*

- $\{\neg\varphi\} \cup \{T \rightarrow\!\!\!» \varphi, T \rightarrow\!\!\!» \psi, \neg\varphi\vee\neg\psi \rightarrow\!\!\!»\neg\varphi\} \Vdash\!\!/\!\approx \psi$

From the perspective of preferential conditional logic, the default T $\rightarrow$» $\psi$ is the only one which may be considered redundant. But then, given $\neg\varphi$, there is no longer any reason to expect $\psi$. Note that this argument holds in particular in a situation where the default $\neg\varphi\vee\neg\psi \rightarrow\!\!\!»\neg\varphi$ has been replaced by the corresponding strict implication and the similarity with **RE** becomes more obvious.

Now, let's see how the defeasible entailment relations mentioned above are handling these "benchmark tests". This will give us a general impression of their overall behaviour and help us to detect major strengths and weaknesses. To compute the inferences for $\Vdash\approx_{\leq,s}$ and $\Vdash\approx_{\leq,p}$, we may use the algorithms described in [Gol 92]. Satisfaction is indicated by **1**, violation by **0**.

| Principles : | Rat° | Pre | Rat | ES | EI | GE | AP | RE | NE |
|---|---|---|---|---|---|---|---|---|---|
| • ⊩≈≤,p | 0 | 1 | 0 | 1 | 1 | 1 | 1 | 1 | 1 |
| • ⊩≈≤,s | 0 | 1 | 1 | 1 | 1 | 1 | 1 | 1 | 1 |
| • ⊩≈≪1 | 1 | 1 | 0 | 1 | 0 | 1 | 1 | 1 | 1 |
| • ⊩≈RC | 1 | 1 | 1 | 1 | 0 | 0 | 0 | 1 | 1 |
| • ⊩≈CE | 0 | 1 | 0 | 1 | 1 | 1 | 0 | 0 | 0 |
| • ⊩≈LC | 0 | 1 | 1 | 1 | 1 | 0 | 0 | 0 | 0 |

All this seems to support our view that nonmonotonic reasoning based on infinitesimal probabilistic knowledge is a natural generalization of classical probabilistic inference and defeasible reasoning with default conditionals. It is particularly well-suited to exploit entropy maximization techniques, offering us very plausible results backed by foundational considerations which are also valid for infinitesimal contexts. Of course, on the other hand, we must see that existing ME-based formalisms are computationally reasonable only or mainly for irredundant default sets, i.e. where no abnormality part A&¬B is covered by the abnormality parts of the remaining defaults. But such a restriction is hardly appealing. However, recent results have indicated that it could be possible to approximate pure ME-default inference by using more practical ranking-based consequence relations admitting nonarchimedean probabilistic justifications. They would allow us to combine rational conditional logic for defaults with a mechanism for exploiting implicit independence assumptions, e.g. to realize **Rat°** and **EI**, which is beyond the scope of the formalisms mentioned above [Wey 95].

## ACKNOWLEDGEMENTS

Thanks to the anonymous referees, who have helped to make this paper a less obscure one.